\pdfoutput=1

\documentclass[11pt]{article}

\usepackage{ACL2023}

\usepackage{times}
\usepackage{latexsym}

\usepackage[T1]{fontenc}

\usepackage[utf8]{inputenc}

\usepackage{microtype}

\usepackage{inconsolata}

\usepackage{csquotes}
\usepackage{amsfonts}
\usepackage{framed}
\usepackage{multirow}

\usepackage{booktabs}

\usepackage{color}
\usepackage{lipsum}
\usepackage{amsmath}
\usepackage{placeins}
\def\parsum#1{\bgroup \textcolor{blue}{Paragraph summary: #1}\egroup}
\def\sectionsum#1{\bgroup \textcolor{green}{Section content: #1}\egroup \\}

\def\parsum#1{}
\def\sectionsum#1{}

\newcommand{\arr}[1]{{#1}}

\title{Private Synthetic Text Generation with Diffusion Models}

\author{
Sebastian Ochs\textsuperscript{$1$}
\and
Ivan Habernal\textsuperscript{$2$} \\
Trustworthy Human Language Technologies \\
\textsuperscript{$1$} Department of Computer Science, Technical University of Darmstadt \\
\textsuperscript{$2$} Research Center Trustworthy Data Science and Security of the University Alliance Ruhr,\\
Faculty of Computer Science, Ruhr University Bochum \\
{\texttt{sebastian.ochs@tu-darmstadt.de, ivan.habernal@ruhr-uni-bochum.de}} \\
\url{www.trusthlt.org}
}

\begin{document}
\maketitle
\begin{abstract}

How capable are diffusion models of generating synthetics texts? Recent research shows their strengths, with performance reaching that of auto-regressive LLMs. 
But are they also good in generating synthetic data if the training was under differential privacy? Here the evidence is missing, yet the promises from private image generation look strong. 
In this paper we address this open question by extensive experiments. 
At the same time, we critically assess (and reimplement) previous works on synthetic private text generation with LLMs and reveal some unmet assumptions that might have led to violating the differential privacy guarantees. 
Our results partly contradict previous non-private findings and show that fully open-source LLMs outperform diffusion models in the privacy regime. 
Our complete source codes, datasets, and experimental setup are publicly available to foster future research\footnote{\url{https://github.com/trusthlt/private-synthetic-text-generation}}.

\end{abstract}

\section{Introduction}
\label{intro}

\parsum{What is known}
How can we share sensitive textual data and protect privacy of individuals in there at the same time?
A go-to method to circumvent this issue is synthetic data generation, used mainly for tabular data \cite{HERNANDEZ202228}, or images \cite{bissoto2018skin} in the medical domain.
However, synthetic data generation alone is not sufficient to protect privacy of the underlying data. For example, \citet{277172} show that outliers in synthetic data suffer from membership inference attacks.

\parsum{What is wrong}
Achieving formal privacy guarantees for the underlying data is possible by combining a generative model with differential privacy (DP, \citet{10.1007/11681878_14}) where we trade off privacy for reduced utility of the synthetic data.
\citet{yue-etal-2023-synthetic} and \citet{mattern-etal-2022-differentially} demonstrated feasibility of synthetic data generation with DP in NLP.
However, both \citet{yue-etal-2023-synthetic} and \citet{mattern-etal-2022-differentially} violate several assumptions about the underlying data which casts doubts on the validity of their findings.

Recent advances also show the success of diffusion models in private synthetic data generation in images \cite{ghalebikesabi2023differentially}. 
Despite recent achievements in conditional text generation using diffusion models \cite{li2022diffusionlm, gong2022diffuseq, 10.5555/3618408.3619275}, the capabilities of private synthetic text generation with diffusion models remain unexplored.

\parsum{What we ask (RQ)}
We thus ask two research questions.
First, what performance on downstream tasks can we achieve using diffusion models for synthetic text generation with varying strengths of differential privacy? 
Second, which factors might have artificially inflated performance in previous works and can we mitigate invalid assumptions in empirical experiments?

\parsum{What we do}
We address these questions as follows. 
Our first hypothesis is that diffusion models for text generation might not suffer from noise added in DP, as they inherently work with a de-noising objective. 
We address this question empirically by conducting extensive (and expensive) experiments with three state-of-the-art diffusion models. 
We address the second research question by two arguments. 
Our first argument is that previous experiments mostly focused on `old' \emph{public} datasets, such as the IMDb movie reviews \cite{maas-etal-2011-learning}. 
As these datasets may very likely have been seen during pretraining of the utilized LLMs (GPT-2), the reported effectiveness of the privacy-preserving synthetic texts may be overestimated. 
We factor out the potential data leaking by introducing five new `fresh' unseen datasets into the experiments. Our second argument is the violation of DP by ignoring group privacy in the datasets. We provide evidence in §\ref{sec:critical.analysis}.

We strive for transparency, reproducibility, and accountability---three key ingredients of privacy-related research. Therefore we experiment with fully open-source and transparent models, such as BLOOM \cite{workshop2023bloom}. All our source codes and datasets are also publicly available for further scrutiny. \arr{Our main contributions are (1) evidence of severe underestimation of DP guarantees in previous works, (2) empirical evidence showing that, unlike in image domain, diffusion models for synthetic text generation suffer severely under DP training, and (3) complete re-implementations of unpublished previous works enabling transparency and potentially increasing trust in privacy-oriented research.}

\section{Theoretical background}

\subsection{Differential privacy (DP)}
DP, introduced by \citet{10.1007/11787006_1}, is a mathematical framework aimed at protecting the privacy of individuals in a dataset.
Through adding a calculated amount of noise to data or statistical queries, it provides formal guarantees that data contributors can not be singled out easily, while still enabling meaningful analysis.
\citet{10.1145/2976749.2978318} apply DP to stochastic gradient descent, called DP-SGD, which allows us to put formal privacy guarantees on trained neural networks.

As the theory of DP is considerably extensive, we refer to \citeposs{wood2018differential} work as an easily accessible introduction for the interested reader.
Furthermore, \citet{habernal-etal-2023-privacy} present a tutorial about how DP can be harnessed for NLP applications, while \citet{Cummings2024Advancing} provide a comprehensive review about the current state of the art in DP research. 

\subsection{Diffusion models for text}
Diffusion models have risen in popularity as generative models recently, especially in the domain of image synthesis \cite{10.1145/3626235}.
Diffusion models can be described as "Markovian Hierarchical Variational Autoencoders" \cite{luo2022understanding}, that are able to generate data from Gaussian noise.
This is achieved by utilizing two transitions during training, called \textit{forward} and \textit{reverse} process.
Given $T$ steps, the \textit{forward} process incrementally adds a small amount of Gaussian noise to a data point $x_0$, creating a chain $x_0, x_1, x_2, ... , x_T$, where $x_T$ resembles a Gaussian distribution.
During the \textit{reverse} process, the diffusion model learns how to transition from $x_i$ to $x_{i-1}, \forall i \in \left \{ 0, 1, 2, ... , T \right \}$, essentially learning how to `undo' the corruption of the \textit{forward} chain, called `denoising'.

For a more mathematically founded description of diffusion models, we point to \citet[chap.~20]{Bishop2024} and the background chapter of \citet{NEURIPS2020_4c5bcfec}.
The \textit{reverse} process can be guided by adding information that relates to the original data.
For example, in text-to-image generation, the latent representation of an image description, using a text encoder, steers the denoising of the diffusion model \cite{ramesh2022hierarchical}.
During inference, the diffusion model is then able to generate images from Gaussian noise that correspond to the input text.

\paragraph{Text diffusion model variants}
Like in image synthesis, diffusion models are also being explored in text generation.
\citet{li2022diffusionlm} propose Diffusion-LM, a diffusion model capable of generating texts non-autoregressively.
In contrast to typical language models, where text sequences are created token-by-token, Diffusion-LM generates a text by gradually denoising a list of Gaussian noise vectors into word embeddings.
The generation process can be directly controlled by providing conditions that the resulting text should fulfill, for example syntactic features such as a predetermined sequence of parts-of-speech tags.

Diffusion-LM's code base also served as a basis for other text diffusion models, such as DiffuSeq \cite{gong2022diffuseq}, a sequence-to-sequence text diffusion model.
Unlike Diffusion-LM, DiffuSeq conditions the generated output on the input text.
This is accomplished through concatenating the word vector sequences of the input and output, and applying the diffusion process only to the output vectors. 
\citet{gong2022diffuseq} report that DiffuSeq achieves similar text-to-text generation capabilities as fine-tuned GPT2-base and -large \cite{radford2019language} models, despite of being non-autoregressive.
Therefore, we included DiffuSeq in our experiments.

Interestingly, while the training and inference method of text diffusion models differ fundamentally from LLMs, their architecture is still based on the Transformer \cite{NIPS2017_3f5ee243}.  

\section{Related work}

To enable the privacy-preserving sharing of labeled text data, \citet{mattern-etal-2022-differentially} propose a method called prompt-based DP fine-tuning, which they utilize to train a GPT2-large model \cite{radford2019language}.
The process can be described as a `reverse' classification task where instead of learning to predict the label for a text, the generative model is trained on generating texts suitable for a given label.
The privacy of each author in the dataset is protected by applying DP-SGD to the fine-tuning procedure.
Additionally, a mismatch loss is applied during training by maximizing the negative log-likelihood of purposely mislabeled texts.
\citet{mattern-etal-2022-differentially} experiment on two publicly available datasets to validate their approach.
The reported results claim that text classifiers trained on the privatized, synthetic data and evaluated on original data experience no significant performance loss compared to a classifier directly trained on the original texts.

\citet{yue-etal-2023-synthetic} use the same methodology in their approach, apart from the mismatch loss.
In contrast to \citet{mattern-etal-2022-differentially}, they include a private customer feedback dataset in their experiments, where the synthetic texts also proved to be useful for downstream task classification performance.
Furthermore, they empirically evaluate the privacy-preservation of the DP generation models by injecting `canaries' into the training data, as suggested by \citet{236216}.
After those models generate a synthetic dataset, it is then possible to track if any canaries have been replicated, which was not the case for models trained with DP guarantees.

Although not in the text domain, diffusion models have also been explored for DP synthetic data generation.
\citet{ghalebikesabi2023differentially} train diffusion models with DP-SGD on several low-resolution image datasets, such as the MNIST dataset of handwritten digits \cite{lecun2010mnist} and generate synthetic images for the downstream classification task.
Classifiers trained on the synthetic data reportedly reach performances close to the state of the art.
Notably, in most experiments, before the diffusion models were trained with DP-SGD on smaller datasets, they were pretrained without DP on the large-scale ImageNet32 \cite{chrabaszcz2017downsampled} dataset.

So far, pretraining large models on public data and fine-tuning those with DP-SGD on smaller, private datasets seems to be an efficient method to produce privacy-preserving and useful synthetic data. 
However, neither Diffusion-LM, nor DiffuSeq, nor most other current diffusion model include pretraining in their methodology.
One exception is GENIE, introduced by \citet{10.5555/3618408.3619275}, which is the only available diffusion model for text-to-text generation that is pretrained. 
GENIE is pretrained on a large corpora of texts by using an objective similar to span-based masked language modeling, 
However, instead of predicting the correct text span, Gaussian noise is continuously added to the selected token span, which GENIE learns to denoise. 
When fine-tuned, the model outperforms the base versions of T5 \cite{10.5555/3455716.3455856} and BART \cite{lewis-etal-2020-bart} in several natural language generation (NLG) tasks. 

\section{Critical analysis of existing works}
\label{sec:critical.analysis}

In the following, we explain in detail how DP synthetic text generation has been accomplished in prior work.
We also critically assess their underlying assumptions and evaluate their respective validity in regards to privacy protection.

\paragraph{Prompting LLMs trained with DP-SGD is correct.}
Both \citet{yue-etal-2023-synthetic} and \citet{mattern-etal-2022-differentially} describe a scenario, where a data holder wishes to benefit from public research on their in-house sensitive text resources (such as medical reports or customer data), but cannot release them due to privacy concerns.
The first assumption is that the sensitive documents are labeled (categorized) and the task to be solved is classification.
The authors address this problem as follows.

First, they create prompts for each original text based on the category to which it belongs.
Second, a pretrained language model is fine-tuned on the prompt:text pairs (e.g., \texttt{<write a positive review: original review text>}) from the training part of the sensitive dataset, learning to create texts resembling the data from the instructed category.
Since synthetic data generation alone is not enough to protect privacy (recall the discussion in Section~\ref{intro}), both authors train their models with DP-SGD.
Afterwards, synthetic texts are sampled from the resulting models using the prompts formulated earlier.
It is worth mentioning that the DP guarantee remains the same regardless of the amount of synthetic generated texts.
The utility of the resulting samples is then evaluated on a downstream classification task, namely by fine-tuning BERT on the generated texts and testing it on the original sensitive test data.

\paragraph{Multiple texts from the same person violates differential privacy.}
\label{sec:assumptions}
The main requirement in DP is that in the underlying dataset, there is a one-to-one correspondence of a person and its data point. From the ML perspective, this means that each example for training or fine-tuning belongs exactly to a unique person. This allows us to spell out the necessary notion of neighboring datasets and the very guarantee of DP, such that the difference of a private analysis will be `roughly the same' (governed by $\varepsilon$ and $\delta$) on two datasets of size $n$ and $n-1$, respectively.

This assumption is violated when a textual dataset contains multiple examples from the same author.
Due to unique writing style, vocabulary or other implicit features, those texts may correlate with each other even though they do not share the same explicit information.
A workaround for this issue is group privacy \citep[Theorem~2.2]{10.1561/0400000042}, which translates to the following:
When assuming that each author provides at most $k$ contributions to a dataset and texts of different authors do not correlate with each other, the privacy bounds of any ($\varepsilon, \delta$)-DP mechanism increase to ($k \varepsilon$, $k \exp((k-1)\varepsilon), \delta$)-DP.
However, to the best of our knowledge, no works utilizing DP-SGD have ever used group privacy, and it is actually unclear whether the DP-SGD is compatible with it.\footnote{Most DP-SGD implementations rely on the amplification theorem by subsampling, so called Poisson sampling, which was only proven for non-group privacy by \citet{Li.et.al.2012.ASIACCS}.} \arr{We offer additional perspectives on this matter in the ``Limitations and ethics statement'' section below.}

Our analysis of the datasets used in previous works reveals that there is a clear violation of this DP assumption. In the book reviews from the Amazon Multi Domain data \cite{blitzer-etal-2007-biographies}, used in \citeposs{mattern-etal-2022-differentially} experiments, users `Shalom Freedman', `Prairie Pal' and 52 more contributed at least more than one review.
Similarly, the Yelp Open Dataset\footnote{\url{https://www.yelp.com/dataset/documentation/main}} provides concrete proof in its documentation that some users have written more than one review. 
Nonetheless, the dataset was part of the experiments carried out by \citet{yue-etal-2023-synthetic}.   

Given these violated assumptions, we cannot really tell whether or not previous works truly guarantee the reported privacy strength. We suspect that under fair conditions the protection would be much lower.

\paragraph{No evidence that the sensitive data were not part of LM pretraining.}
Another assumption in \citeposs{mattern-etal-2022-differentially} and \citeposs{yue-etal-2023-synthetic} work is that the `sensitive' data (in this case IMDb, Yelp, etc.) have not been `seen' during pretraining of the generative language model.
We identify two potential issues here.

First, any data point accessed during LLM pretraining can be potentially leaked by adversarial prompting \citep{274574,nasr2023scalable}. If the sensitive data were used both in (1) LLM pretraining and in (2) private fine-tuning, privacy had been breached in (1) already and no claims about protection in (2) are valid.

Second, pre-training and fine-tuning on the same data will most likely boost the effectiveness of the synthesized texts, as opposed to synthesizing out-of-domain `fresh' data. This has been demonstrated in previous work by \citet{igamberdiev-etal-2022-dp} who found that such leaking led to unrealistically good results in other works.

Since \citet{yue-etal-2023-synthetic} and \citet{mattern-etal-2022-differentially}  use GPT2 \cite{radford2019language} in their method, and the pretraining data of that model is not disclosed, we doubt that the performance reported on the YELP and IMDb dataset is fully transferable.
Hence, in our experiments we only include datasets that are \emph{guaranteed} not to be part of the pretraining data.

\paragraph{Lack of transparency and reproducibility.}
Despite claims to publish their code, \citet{mattern-etal-2022-differentially} have yet to follow through on this promise (requests from our side to gain access to their source code have unfortunately been evaded).\footnote{\url{https://github.com/justusmattern/private-datasets-with-llms}}
For the sake of reproducibility, we therefore partly implement their approach based on the information provided in their paper.
We decide against utilizing the proposed mismatch loss in our experiments in Section \ref{sec:experiments}, as the increase in downstream task performance was disproportionally low compared to the increase in computation time and memory usage.
Moreover, we publicly release our implemented version of \citeposs{mattern-etal-2022-differentially} work to make it available for the NLP community (please refer to the submission supplementary materials for now).

\section{Experiments}
\label{sec:experiments}
We investigate the capabilities of diffusion models to generate synthetic texts under DP with the following experimental setup.

\subsection{Why diffusion for private text generation?} \label{diffuser}
When training a neural network with DP-SGD, the gradients of the model are modified through gradient clipping and adding Gaussian noise, to prevent it from learning "too much" about a given data sample.
In the case of synthetic text generation with LLMs, this may lead to incoherent text samples, as described by \citet{mattern-etal-2022-differentially}.
In the image domain, however, \citet{dockhorn2023differentially} argue that the denoising module of diffusion models reacts less sensitive to the gradient modifications caused by DP-SGD, at least in comparison to a generative adversarial network (GAN). 
This claim is supported by the promising results that both \citet{dockhorn2023differentially} and \citet{ghalebikesabi2023differentially} accomplished for their DP synthetic image synthesis approaches.
We therefore hypothesize that text diffusion models, due to their unique training and inference methodology, may be more resilient towards the noise introduced by DP than their LLM counterparts.
However, to the best of our knowledge, text diffusion models have not yet been explored for privacy-preserving methods.
For this reason, we compare the non-autoregressive diffusion models to the conventional token-by-token producing LLMs for DP synthetic text generation.

\subsection{Data}
As we discussed the disadvantages of using data that potentially has been encountered during pretraining, we do not experiment on any of the datasets from previous works.
Instead, we train our models on the following four datasets.

First, we include a collection of non-spam and spam emails in our experiments, published by \citet{alsubaiey2024novelinterpretablerobustwebbased}, termed SPAM dataset in this work.
It combines several spam detection datasets into one corpus of 82k mails (66k train, 16k test; 43k spam, 40k "ham").
The data is interesting from a privacy sensitive viewpoint, as it also contains company-internal mails from the Enron email dataset \cite{enron}. 
While standard text classifiers already achieve high performances on this SPAM dataset, it is challenging for synthetic text generation, due to the noisiness of the spam emails.

Second, we incorporate a dataset addressing mental health issues in our research, specifically the "Reddit SuicideWatch and Mental Health Collection" \cite{Ji2021}, and referred to as SWMH.
The 54k (35k train, 9k validation, 11k test) data samples were collected from five subreddits revolving around mental health challenges and suicidal ideation, often containing personal details about the life of the respective authors.
Consequently, the access to this data is restricted to research purposes only.
As each post is labeled by the name of its original forum (subreddit), the dataset is suitable for text classification.

\label{thumbs-up}
Third, we also add the "Thumbs Up" dataset \cite{10020586} to our experimental suite.  
It consists of 2.1 million negative Google Play Store app reviews (1.2M train, 420k validation, 570k test), collected between October 2021 and March 2022.
Each review is annotated based on the number of upvotes received from other users within a month.
According to \citet{10020586}, a high number of upvotes may indicate an app issue that multiple users encountered, while reviews with few upvotes may be less relevant for app improvement.  
The reviews are segmented into five classes (0 votes, 1-5 votes, 6-25 votes, 26-100 votes, >100 votes).
We estimate that the data is challenging for synthetic text generation due to two reasons: First, the label distribution is highly skewed towards the lower classes. 
Second, a BERT classifier trained on the original training data and evaluated on the development set only achieved an accuracy of $0.66$ and a macro F1-score of $0.34$. 
More importantly, the app reviews have been created after the release of some popular LLMs, which automatically excludes them from the pretraining data of GPT2.

Last, but not least, we rely on a collection of drug reviews from WebMD\footnote{\url{https://www.webmd.com/}} for our experiments, published by \citet{Harode2020}.
The web scrape provides 363k detailed user experiences with medical drugs: each review is accompanied by the the user's age, gender and medical condition, the name of the drug, and user ratings of satisfaction, ease of use and drug effectiveness.
Similar to the SWMH dataset, the review texts may reveal personal information, which necessitates privacy-preserving methods.
In our experiments, we use the satisfaction rating for the text classification task.

Unfortunately, neither dataset discloses authorship information such as user names or unique identification numbers, preventing us to create a one-to-one relation between a review and its author. 
Therefore some of the critique from Section~\ref{sec:assumptions} can be raised against our main experiments on the four adopted datasets; we are aware of this issue. 
Here we had to trade off accessibility and size (i.e., Thumbs Up being a large and publicly available dataset) for strict privacy (such as working on truly private in-house data which could not be shared with the public).
\arr{However, to overcome this inherent problem in existing datasets, we created an additional fifth dataset from scratch similar to the WebMD datasets originating from \texttt{drugs.com}, to ensure a one-to-one relation between authors and their texts. We report on this additional experiment in Appendix \ref{supp}.}

\subsection{Tasks}
We model our approach for synthetic text generation after the method proposed by \citet{mattern-etal-2022-differentially}.
Both the SPAM and SWMH dataset are fairly balanced, so we utilize their provided training sets.
As the Thumbs-Up dataset is biased towards app reviews with no upvotes, we balance the training set so that each label is represented by an equal amount of samples. 
We also balance and split the WebMD drug reviews into a training (80\%) and test set (20\%).
A detailed overview of all data splits is provided in the Appendix Section \ref{data-dist}.
 
Guided by the label, instructions are created for each text, following the format: \\ \newline
\textbf{SPAM:} "write a (spam | non-spam) e-mail:" \\ 
\textbf{SWMH:} "write a post to the (anxiety | bipolar | depression | offmychest | suicidewatch) community:" \\ 
\textbf{Thumbs-Up:} "write a (mild | notable | concerning | serious | hot) negative app review: " \\ 
\textbf{WebMD:} "write a (terrible | poor | neutral | good | great) medicine review: " \\ \newline
We privately (DP-SGD) fine-tune each language model in our experiments using the instructions as input and the corresponding review as output.
After training, we sample 1,000 synthetic texts from each model, again with an equal distribution of labels.
We evaluate the utility of the synthetic data with a BERT \cite{devlin-etal-2019-bert} classifier.
The classifier is trained on the synthetic texts for 5 epochs (learning rate = $2e-5$), aimed at predicting the label corresponding to the instruction each sample was generated with.
We validate it after each epoch on either a subsample (2k) of the validation set (Thumbs Up, SWMH) or the original train data (SPAM, WebMD) and select the best classifier in terms of macro F1-score.
We then measure the performance of each classifier on the original test sets, based on accuracy and macro F1-score, and evaluate the quality of the synthetic texts with the perplexity score, using BLOOM-560m \cite{workshop2023bloom} as base model.

\subsection{Diffusion models}
For our experiments, we consider three text diffusion models, DiffuSeq \cite{gong2022diffuseq}, SeqDiffuSeq \cite{yuan-etal-2024-text} and GENIE \cite{10.5555/3618408.3619275}.

In contrast to other transformer-based language models, most diffusion models are not pretrained.
While this does not seem to impact model performance when training under non-private settings, preliminary experiments with DiffuSeq demonstrated that training the conditional diffusion model from scratch with DP-SGD introduced too much noise, and it was unable to generate any text.
Nonetheless, to enable private training, we use a public checkpoint of DiffuSeq trained on the Quora Question Pair (QQP) dataset \footnote{\url{https://www.kaggle.com/c/quora-question-pairs}} and fine-tune it with DP-SGD on our selected data collections.

SeqDiffuSeq is an encoder-decoder-based diffusion model. 
It achieves better text generation capabilities than DiffuSeq due to using a self-conditioning method from the image domain \cite{chen2023analogbitsgeneratingdiscrete} and a token-based adaptive noise schedule. 
We also mitigate the noise introduced by DP-SGD by pretraining SeqDiffuSeq on the commonsense dialogue dataset, following the settings from \citet{yuan-etal-2024-text}.
 
Another model representing diffusion models in our experiments is GENIE, the only classically pretrained text-to-text diffusion model to our knowledge.  
GENIE is pretrained on the same text corpus as BART \cite{lewis-etal-2020-bart} and uses a span-based masked language modeling objective.
However, rather than replacing the spans with mask tokens, the forward diffusion process is applied to the spans instead.

\begin{table*}[!t]
\begin{center}
\begin{small}
\begin{tabular}{p{2cm}rrrrrrrrr} \toprule
\textbf{SPAM} & \multicolumn{3}{c}{$\varepsilon = 3$} & \multicolumn{3}{c}{$\varepsilon = 8$} & \multicolumn{3}{c}{$\varepsilon = \infty$} \\
& Acc & MF1 & PPL & Acc & MF1 & PPL & Acc & MF1 & PPL\\ \midrule
\multicolumn{1}{l|}{BART} & 0.77 & \textbf{0.77} & \multicolumn{1}{c|}{55} & 0.82  & 0.81 & \multicolumn{1}{c|}{192} & 0.81 & 0.81            & \multicolumn{1}{c}{\textbf{133}}  \\
\multicolumn{1}{l|}{BLOOM}  & 0.55 & 0.44 & \multicolumn{1}{c|}{55} & 0.49  & 0.47 & \multicolumn{1}{c|}{61}          & \textbf{0.94} & \textbf{0.94}            & \multicolumn{1}{c}{165}  \\
\multicolumn{1}{l|}{PHI-1.5}   & \textbf{0.78} & \textbf{0.77} & \multicolumn{1}{c|}{\textbf{40}} & \textbf{0.85}  & \textbf{0.85} & \multicolumn{1}{c|}{\textbf{45}}          & 0.80 & 0.79            & \multicolumn{1}{c}{230}  \\ \midrule
\multicolumn{1}{l|}{DIFFUSEQ}   & 0.45 & 0.31 & \multicolumn{1}{c|}{-} & 0.55  & 0.35 & \multicolumn{1}{c|}{-}          & \textbf{0.94} & 0.93            & \multicolumn{1}{c}{1686}  \\
\multicolumn{1}{l|}{SEQDIFFUSEQ}   & 0.58 & 0.49 & \multicolumn{1}{c|}{454} & 0.57  & 0.57 & \multicolumn{1}{c|}{330}          & 0.57 & 0.43            & \multicolumn{1}{c}{2547}  \\
\multicolumn{1}{l|}{GENIE}   & 0.63 & 0.63 & \multicolumn{1}{c|}{8e+10} & 0.53  & 0.50 & \multicolumn{1}{c|}{2e+4}          & 0.83 & 0.82            & \multicolumn{1}{c}{2e+5}  \\ \toprule
\textbf{SWMH} & \multicolumn{3}{c}{$\varepsilon = 3$} & \multicolumn{3}{c}{$\varepsilon = 8$} & \multicolumn{3}{c}{$\varepsilon = \infty$} \\
& Acc & MF1 & PPL & Acc & MF1 & PPL & Acc & MF1 & PPL\\ \midrule
\multicolumn{1}{l|}{BART} & 0.29 & 0.29 & \multicolumn{1}{c|}{1184} & 0.32  & 0.33 & \multicolumn{1}{c|}{\textbf{10}} & 0.23 & 0.20            & \multicolumn{1}{c}{102}  \\
\multicolumn{1}{l|}{BLOOM}  & 0.34 & 0.32 & \multicolumn{1}{c|}{3e+4} & \textbf{0.49}  & 0.42 & \multicolumn{1}{c|}{819}          & \textbf{0.58} & \textbf{0.57}            & \multicolumn{1}{c}{\textbf{32}}  \\
\multicolumn{1}{l|}{PHI-1.5}   & 0.26 & 0.24 & \multicolumn{1}{c|}{\textbf{24}} & 0.46  & \textbf{0.48} & \multicolumn{1}{c|}{26}          & 0.43 & 0.47            & \multicolumn{1}{c}{38}  \\ \midrule
\multicolumn{1}{l|}{DIFFUSEQ}   & - & - & \multicolumn{1}{c|}{-} & -  & - & \multicolumn{1}{c|}{-}          & 0.50 & 0.45            & \multicolumn{1}{c}{475}  \\
\multicolumn{1}{l|}{SEQDIFFUSEQ}   & 0.21 & 0.17 & \multicolumn{1}{c|}{9e+11} & 0.34  & 0.19 & \multicolumn{1}{c|}{2e+7}          & 0.31 & 0.26            & \multicolumn{1}{c}{1+e12}  \\
\multicolumn{1}{l|}{GENIE}   & \textbf{0.40} & \textbf{0.40} & \multicolumn{1}{c|}{2e+5} & 0.43  & 0.41 & \multicolumn{1}{c|}{3e+5}          & 0.39 & 0.37            & \multicolumn{1}{c}{2e+5}  \\ \toprule
\textbf{THUMBSUP} & \multicolumn{3}{c}{$\varepsilon = 3$} & \multicolumn{3}{c}{$\varepsilon = 8$} & \multicolumn{3}{c}{$\varepsilon = \infty$} \\
& Acc & MF1 & PPL & Acc & MF1 & PPL & Acc & MF1 & PPL\\ \midrule
\multicolumn{1}{l|}{BART} & 0.21 & 0.14 & \multicolumn{1}{c|}{7e+5} & 0.18  & 0.15 & \multicolumn{1}{c|}{1e+5} & 0.15 & 0.14 & \multicolumn{1}{c}{\textbf{20}}  \\
\multicolumn{1}{l|}{BLOOM}  & 0.22 & 0.13 & \multicolumn{1}{c|}{176} & 0.21  & 0.17 & \multicolumn{1}{c|}{119}          & 0.22 & 0.16            & \multicolumn{1}{c}{42}  \\
\multicolumn{1}{l|}{PHI-1.5}   & 0.21 & 0.18 & \multicolumn{1}{c|}{\textbf{35}} & 0.21  & 0.14 & \multicolumn{1}{c|}{\textbf{35}}          & 0.20 & 0.18            & \multicolumn{1}{c}{119}  \\ \midrule
\multicolumn{1}{l|}{DIFFUSEQ}   & 0.20 & 0.15 & \multicolumn{1}{c|}{-} & 0.20  & 0.09 & \multicolumn{1}{c|}{-}          & 0.30 & 0.28            & \multicolumn{1}{c}{9e+13}  \\
\multicolumn{1}{l|}{SEQDIFFUSEQ}   & 0.23 & 0.22 & \multicolumn{1}{c|}{7e+12} & 0.17  & 0.16 & \multicolumn{1}{c|}{4e+12}          & 0.20 & 0.07            & \multicolumn{1}{c}{3e+6}  \\
\multicolumn{1}{l|}{GENIE}   & 0.24 & 0.19 & \multicolumn{1}{c|}{2e+13} & 0.27  & 0.23 & \multicolumn{1}{c|}{6e+11}          & 0.36 & 0.33            & \multicolumn{1}{c}{1e+13}  \\ \toprule
\textbf{WEBMD} & \multicolumn{3}{c}{$\varepsilon = 3$} & \multicolumn{3}{c}{$\varepsilon = 8$} & \multicolumn{3}{c}{$\varepsilon = \infty$} \\
& Acc & MF1 & PPL & Acc & MF1 & PPL & Acc & MF1 & PPL\\ \midrule
\multicolumn{1}{l|}{BART} & 0.22 & 0.12 & \multicolumn{1}{c|}{3e+13} & 0.21  & 0.14 & \multicolumn{1}{c|}{3e+14} & 0.23 & 0.17            & \multicolumn{1}{c}{2e+14}  \\
\multicolumn{1}{l|}{BLOOM}  & 0.19 & 0.17 & \multicolumn{1}{c|}{85} & 0.21  & 0.15 & \multicolumn{1}{c|}{66}          & 0.22 & 0.21            & \multicolumn{1}{c}{\textbf{37}}  \\
\multicolumn{1}{l|}{PHI-1.5}   & 0.21 & 0.11 & \multicolumn{1}{c|}{\textbf{26}} & 0.20  & 0.11 & \multicolumn{1}{c|}{\textbf{26}}          & 0.21 & 0.17            & \multicolumn{1}{c}{60}  \\ \midrule
\multicolumn{1}{l|}{DIFFUSEQ}   & 0.20 & 0.07 & \multicolumn{1}{c|}{-} & 0.20  & 0.14 & \multicolumn{1}{c|}{-}          & 0.22 & 0.20            & \multicolumn{1}{c}{9e+12}  \\
\multicolumn{1}{l|}{SEQDIFFUSEQ}   & 0.21 & 0.16 & \multicolumn{1}{c|}{1e+11} & 0.20  & 0.19 & \multicolumn{1}{c|}{5e+5}          & 0.30 & 0.26            & \multicolumn{1}{c}{4e+6}  \\
\multicolumn{1}{l|}{GENIE}   & 0.22 & 0.17 & \multicolumn{1}{c|}{1e+13} & 0.22  & 0.15 & \multicolumn{1}{c|}{5e+12}          & 0.21 & 0.19            & \multicolumn{1}{c}{4e+12}  \\
\end{tabular}
\end{small}
\end{center}
\caption{\label{expresults} Accuracy (Acc) and macro F1-score (MF1) results of the BERT classifier trained on synthetic data and tested on the original test sets. We also display the average perplexity (PPL) of the synthetic data. For each $\varepsilon$ value, we highlight the best Acc($\uparrow$) and MF1($\uparrow$) score for the SPAM and SWMH dataset, while the best PPL($\downarrow$) score is emphasized across all datasets.}
\end{table*}

\subsection{Baselines}
Albeit both \citet{mattern-etal-2022-differentially} and \citet{yue-etal-2023-synthetic}  use GPT2-XL \cite{radford2019language}, we do not include it as baseline in our experiments to avoid potential pretraining data leakage. 
Since GENIE and BART \cite{lewis-etal-2020-bart} share the same pretraining data, we add BART-large as second baseline. 
Additionally, BARTs pretrained checkpoints were published before the "Thumbs Up" dataset.

Another large language model we utilize as baseline is BLOOM \cite{workshop2023bloom}.
Even though it released after the data collection of "Thumbs Up", due to the transparency of BLOOM and the option to scour its corpus with the ROOTS search tool \cite{piktus-etal-2023-roots}, we are able to conclude that none of our selected datasets are present in its pretraining data.
We ensure that BLOOM is of similar size in regards to BART-large by selecting the 560 million parameter version of BLOOM as another baseline.

Our final baseline is Phi-1.5 \cite{li2023textbooksneediiphi15}, a 1.3 billion transformer-based model, pretrained on synthetically generated text data, which ensures that it has not encountered any of our chosen datasets before fine-tuning.    

\subsection{Fine-Tuning}
All models were fine-tuned with an 80 GB NVIDIA A100 Tensor Core GPU.
Utilizing the Opacus \cite{opacus} library and with some modifications, we are able to train all baselines and diffusion models on our data with DP-SGD.
For the interested reader, we provide a detailed description of our implementation in the Appendix \ref{dp-sgd-section}. 
As for our privacy budgets, we decide on $\varepsilon = 3$ and $\varepsilon = 8$, using \citet{mattern-etal-2022-differentially}'s work as reference. 
However, we select a stronger $\delta = (10 * \# \textit{ of training samples}) ^{-1}$ to effectively reduce the chance of accidentally leaking any private information of a single individual.
In addition, we also fine-tune all models without DP, to measure how much utility is lost when protecting privacy. 
For detailed hyperparameter settings, please refer to the Appendix Section \ref{sec:hyperparams}.
In general, training BLOOM and PHI-1.5 with DP-SGD on our largest dataset takes 1.5 days, while DP finetuning DiffuSeq takes 3 days.   

\subsection{Results}
The experimental results are presented in Table~\ref{expresults}. 
We also offer randomly sampled texts from all models as well as the classifier performance on the original data in Appendix \ref{moresamples}.

For the SPAM and SWMH dataset, non-DP training usually outperforms all DP approaches.
The utility drop-off when using DP varies often, impacting BLOOMs performance negatively on both SPAM and SWMH, while BART and Phi-1.5 are almost unaffected on SPAM.
All text diffusion models are usually outclassed by their LLM baselines on the SPAM dataset, except for DiffuSeq in the non-DP setting.
\arr{This is also supported by our supplementary experiment in Appendix \ref{supp}, where GENIE is surpassed by BART and PHI-1.5 under DP constraints.}
The SWMH dataset displays a more leveled field, albeit BLOOM mostly achieves better scores than the diffusion models.
 
The metric results for the Thumbs-Up and WebMD dataset indicate that these tasks are more challenging, as all classifiers perform close to random chance when DP is applied. 
We even observe higher metrics for non-DP DiffuSeq and GENIE on the Thumbs-Up dataset, while SeqDiffuSeq performs slightly better than all other models on WebMD.   

When considering perplexity, Phi-1.5 often displays lower across all settings than all other models, while the text diffusion models have orders of magnitudes higher perplexity scores than the LLM baselines.

Overall, the results do not meet the expectations.
In Section \ref{diffuser}, we hypothesized that the diffusion models are less impacted by the noise introduced by DP-SGD due to their unique training and inference based on Gaussian noise. 
On the contrary, when using DP, DiffuSeq often repeats a single word over and over again, and does not generate any text for the SWMH dataset.
Reading the texts generated by the text diffusion models also reveals that they are often incoherent, even in the non-private setting, as estimated by their high perplexity scores. 

\section{Discussion}
\label{discuss}
In contrast to our experiments, the samples provided in \citet[Table 9]{10.5555/3618408.3619275} demonstrate that diffusion models are capable of generating semantically and syntactically correct texts.
This may be mainly caused by the task differences between our works.
\citet[Table 9]{10.5555/3618408.3619275}, \citet[Table 1]{gong2022diffuseq} and \citet[Table 1]{yuan-etal-2024-text} fine-tune their models on single-sentence datasets, e.g. XSUM \cite{narayan-etal-2018-dont} for GENIE:
The average sequence output length of this particular text summarization dataset is significantly lower ($19.77$ words ) than those of our utilized SPAM ($264.15$ words), SWMH ($208.05$ words) and WebMD ($62.00$ words) datasets.
Consequently, a manual comparison of the generated texts (Appendix \ref{moresamples}) shows that the LLMs are able to generate coherent texts even under DP conditions, while the diffusion models struggle to do so.
Another advantage our LLM baselines have over diffusion models is inference speed, as they generate 1,000 text samples on average $36$ times faster than GENIE, and $250$ times faster than DiffuSeq.

Albeit the experimental results of combining DP with text diffusion models are not promising, we still encourage the exploration of this method in future work.
We believe that major improvements to this model class are possible, as demonstrated by \citet{gong-etal-2023-diffuseq}, which drastically improved the training and inference speed of DiffuSeq.

For privacy-preserving NLP research involving DP, we recommend to take Section \ref{sec:assumptions} into consideration for their experiments.
Future research should not fine-tune LLMs with DP-SGD without justifying how the main requirements of DP are met in their work. 

\section{Conclusion}
We explored the capability of three text-to-text diffusion models to generate private synthetic texts.
We also revealed that reported DP guarantees are severely underestimated by previous works for synthetic text generation, due to omitting group privacy rules and potential leakage of the employed datasets into the pretraining data of the involved LLMs.
While our experiments demonstrate that diffusion models do not exceed their baselines for private synthetic text generation, we are the first to train a text diffusion model with DP constraints.
Hopefully, this enables further exploration of privacy methods incorporating these type of models.

\section*{Limitations and ethics statement}
While fine-tuning strategies for LLMs are widely explored, the same is not the case for the just upcoming text diffusion models.
Especially when DP-SGD is involved, it can be challenging to select a working hyperparameter setting.
We therefore cannot guarantee that we utilized the optimal hyperparameters for our diffusion models, which may have influenced the experimental results unfavorably.

Unfortunately, due to high computation costs of training text diffusion models and slow inference time (1,000 samples from DiffuSeq took almost two days), we were unable to generate larger quantities of synthetic texts and are limited to 1,000 samples per synthetic dataset, which may lower their viability as replacement for the original data.

The underlying dependency of the "Thumbs Up" data labels on user upvotes makes it more challenging for training any classifier, especially considering that the distribution of upvotes onto negative app reviews addressing the same issue can highly vary. 
As the "Thumbs Up" dataset also provides the exact amount of user upvotes for each review, exploring the task as regression might have been more suitable. 

To the best of our knowledge, no published text classification research has utilized the WebMD and SWMH dataset, which is why we default to the standard classification procedure for NLP. A more sophisticated text classification approach may have resulted in stronger experimental results, however, this was not the main focus of our research.

Although the synthetic texts generated by our models trained on SWMH are particularly interesting, we decided against sharing them publicly, as the dataset has a restrictive access policy, and, as we stated in our critique of previous works in Section~\ref{sec:assumptions}, the provided privacy guarantees by our own experiments may also be overestimated.

As mentioned in Section \ref{discuss}, the original experiments of DiffuSeq and SeqDiffuSeq explore text generation tasks with target sequences much shorter than found in our selected datasets. 
While their architectures allow output sequences of up to 256 tokens, their generation capabilities for texts beyond the length of a single sentence are underexplored.
GENIE, on the other hand, was pretrained with target sequences with up to 153 tokens, and did not perform significantly better than its text diffusion peers in the measured utility metrics.
It could be possible to decrease the noise introduced through DP-SGD by increasing the number of diffusion steps beyond 2000. We left this value untouched to not further increase inference time, considering our computational budget, but it should be worthwhile to explore it in future work.

Our view on the one-to-one correspondence of a person and the data sample is very strict but it is based on our understanding of the foundations of DP. For example, \citet[p.~130]{Blum.et.al.2005} clearly stated the underlying assumptions about a datasets to be processed by a DP analysis---it is a table where each row corresponds to a person. Thus having a dataset where multiple rows (ie.\ samples) belong to one person violates this assumption. However, the NLP research somehow implicitly moved away from this assumption (which has not been much emphasized in the later literature on NLP/ML with DP) and proposed, for example, document-level privacy. Without debating the underlying philosophy what privacy is, in our view what we DP really cares for is privacy of entities (e.g., persons, groups) and not objects (e.g., documents, words). Some recent examples from different communities also discuss the privacy notion very explicitly, e.g., protecting users in Wikipedia edits at TPDP'23 \citep{adeleye2023publishing}.

\section*{Acknowledgements}
This work has been partly supported by the Research Center Trustworthy Data
Science and Security (\url{https://rc-trust.ai}), one of the Research Alliance
centers within the \url{https://uaruhr.de} 
and by the German Federal Ministry of Education
and Research and the Hessian Ministry of Higher
Education, Research, Science and the Arts within
their joint support of the National Research Center
for Applied Cybersecurity ATHENE.

\bibliography{custom}

\begin{thebibliography}{53}
\expandafter\ifx\csname natexlab\endcsname\relax\def\natexlab#1{#1}\fi

\bibitem[{Abadi et~al.(2016)Abadi, Chu, Goodfellow, McMahan, Mironov, Talwar, and Zhang}]{10.1145/2976749.2978318}
Martin Abadi, Andy Chu, Ian Goodfellow, H.~Brendan McMahan, Ilya Mironov, Kunal Talwar, and Li~Zhang. 2016.
\newblock \href {https://doi.org/10.1145/2976749.2978318} {{Deep Learning with Differential Privacy}}.
\newblock In \emph{Proceedings of the 2016 ACM SIGSAC Conference on Computer and Communications Security}, CCS '16, page 308–318, New York, NY, USA. Association for Computing Machinery.

\bibitem[{Adeleye et~al.(2023)Adeleye, Berghel, Desfontaines, Hay, Johnson, Lemoisson, Machanavajjhala, Magerlein, Modena, Pujol et~al.}]{adeleye2023publishing}
Temilola Adeleye, Skye Berghel, Damien Desfontaines, Michael Hay, Isaac Johnson, Cl{\'e}o Lemoisson, Ashwin Machanavajjhala, Tom Magerlein, Gabriele Modena, David Pujol, et~al. 2023.
\newblock Publishing wikipedia usage data with strong privacy guarantees.
\newblock \emph{arXiv preprint}.

\bibitem[{Al-Subaiey et~al.(2024)Al-Subaiey, Al-Thani, Alam, Antora, Khandakar, and Zaman}]{alsubaiey2024novelinterpretablerobustwebbased}
Abdulla Al-Subaiey, Mohammed Al-Thani, Naser~Abdullah Alam, Kaniz~Fatema Antora, Amith Khandakar, and SM~Ashfaq~Uz Zaman. 2024.
\newblock \href {http://arxiv.org/abs/2405.11619} {Novel interpretable and robust web-based ai platform for phishing email detection}.

\bibitem[{Ansel et~al.(2024)Ansel, Yang, He, Gimelshein, Jain, Voznesensky, Bao, Bell, Berard, Burovski, Chauhan, Chourdia, Constable, Desmaison, DeVito, Ellison, Feng, Gong, Gschwind, Hirsh, Huang, Kalambarkar, Kirsch, Lazos, Lezcano, Liang, Liang, Lu, Luk, Maher, Pan, Puhrsch, Reso, Saroufim, Siraichi, Suk, Zhang, Suo, Tillet, Zhao, Wang, Zhou, Zou, Wang, Mathews, Wen, Chanan, Wu, and Chintala}]{pytorch}
Jason Ansel, Edward Yang, Horace He, Natalia Gimelshein, Animesh Jain, Michael Voznesensky, Bin Bao, Peter Bell, David Berard, Evgeni Burovski, Geeta Chauhan, Anjali Chourdia, Will Constable, Alban Desmaison, Zachary DeVito, Elias Ellison, Will Feng, Jiong Gong, Michael Gschwind, Brian Hirsh, Sherlock Huang, Kshiteej Kalambarkar, Laurent Kirsch, Michael Lazos, Mario Lezcano, Yanbo Liang, Jason Liang, Yinghai Lu, C.~K. Luk, Bert Maher, Yunjie Pan, Christian Puhrsch, Matthias Reso, Mark Saroufim, Marcos~Yukio Siraichi, Helen Suk, Shunting Zhang, Michael Suo, Phil Tillet, Xu~Zhao, Eikan Wang, Keren Zhou, Richard Zou, Xiaodong Wang, Ajit Mathews, William Wen, Gregory Chanan, Peng Wu, and Soumith Chintala. 2024.
\newblock \href {https://doi.org/10.1145/3620665.3640366} {Pytorch 2: Faster machine learning through dynamic python bytecode transformation and graph compilation}.
\newblock In \emph{Proceedings of the 29th ACM International Conference on Architectural Support for Programming Languages and Operating Systems, Volume 2}, ASPLOS ’24. ACM.

\bibitem[{Bishop and Bishop(2024)}]{Bishop2024}
Christopher~M. Bishop and Hugh Bishop. 2024.
\newblock \href {https://doi.org/10.1007/978-3-031-45468-4} {\emph{{Deep Learning: Foundations and Concepts}}}.
\newblock Springer International Publishing.

\bibitem[{Bissoto et~al.(2018)Bissoto, Perez, Valle, and Avila}]{bissoto2018skin}
Alceu Bissoto, F\'abio Perez, Eduardo Valle, and Sandra Avila. 2018.
\newblock {Skin Lesion Synthesis with Generative Adversarial Networks}.
\newblock In \emph{OR 2.0 Context-Aware Operating Theaters, Computer Assisted Robotic Endoscopy, Clinical Image-Based Procedures, and Skin Image Analysis}, pages 294--302. Springer.

\bibitem[{Blitzer et~al.(2007)Blitzer, Dredze, and Pereira}]{blitzer-etal-2007-biographies}
John Blitzer, Mark Dredze, and Fernando Pereira. 2007.
\newblock \href {https://aclanthology.org/P07-1056} {Biographies, {B}ollywood, boom-boxes and blenders: Domain adaptation for sentiment classification}.
\newblock In \emph{Proceedings of the 45th Annual Meeting of the Association of Computational Linguistics}, pages 440--447, Prague, Czech Republic. Association for Computational Linguistics.

\bibitem[{Blum et~al.(2005)Blum, Dwork, McSherry, and Nissim}]{Blum.et.al.2005}
Avrim Blum, Cynthia Dwork, Frank McSherry, and Kobbi Nissim. 2005.
\newblock \href {https://doi.org/10.1145/1065167.1065184} {{Practical privacy: the SuLQ framework}}.
\newblock In \emph{Proceedings of the Twenty-Fourth ACM SIGMOD-SIGACT-SIGART Symposium on Principles of Database Systems}, PODS '05, page 128–138, New York, NY, USA. Association for Computing Machinery.

\bibitem[{Carlini et~al.(2019)Carlini, Liu, Erlingsson, Kos, and Song}]{236216}
Nicholas Carlini, Chang Liu, {\'U}lfar Erlingsson, Jernej Kos, and Dawn Song. 2019.
\newblock \href {https://www.usenix.org/conference/usenixsecurity19/presentation/carlini} {{The Secret Sharer: Evaluating and Testing Unintended Memorization in Neural Networks}}.
\newblock In \emph{28th USENIX Security Symposium (USENIX Security 19)}, pages 267--284, Santa Clara, CA. USENIX Association.

\bibitem[{Carlini et~al.(2021)Carlini, Tram{\`e}r, Wallace, Jagielski, Herbert-Voss, Lee, Roberts, Brown, Song, Erlingsson, Oprea, and Raffel}]{274574}
Nicholas Carlini, Florian Tram{\`e}r, Eric Wallace, Matthew Jagielski, Ariel Herbert-Voss, Katherine Lee, Adam Roberts, Tom Brown, Dawn Song, {\'U}lfar Erlingsson, Alina Oprea, and Colin Raffel. 2021.
\newblock \href {https://www.usenix.org/conference/usenixsecurity21/presentation/carlini-extracting} {{Extracting Training Data from Large Language Models}}.
\newblock In \emph{30th USENIX Security Symposium (USENIX Security 21)}, pages 2633--2650. USENIX Association.

\bibitem[{Chen et~al.(2023)Chen, Zhang, and Hinton}]{chen2023analogbitsgeneratingdiscrete}
Ting Chen, Ruixiang Zhang, and Geoffrey Hinton. 2023.
\newblock \href {http://arxiv.org/abs/2208.04202} {Analog bits: Generating discrete data using diffusion models with self-conditioning}.

\bibitem[{Chrabaszcz et~al.(2017)Chrabaszcz, Loshchilov, and Hutter}]{chrabaszcz2017downsampled}
Patryk Chrabaszcz, Ilya Loshchilov, and Frank Hutter. 2017.
\newblock \href {https://doi.org/10.48550/arXiv.1707.08819} {{A Downsampled Variant of ImageNet as an Alternative to the CIFAR datasets}}.
\newblock \emph{arXiv preprint}.

\bibitem[{Cummings et~al.(2024)Cummings, Desfontaines, Evans, Geambasu, Huang, Jagielski, Kairouz, Kamath, Oh, Ohrimenko, Papernot, Rogers, Shen, Song, Su, Terzis, Thakurta, Vassilvitskii, Wang, Xiong, Yekhanin, Yu, Zhang, and Zhang}]{Cummings2024Advancing}
Rachel Cummings, Damien Desfontaines, David Evans, Roxana Geambasu, Yangsibo Huang, Matthew Jagielski, Peter Kairouz, Gautam Kamath, Sewoong Oh, Olga Ohrimenko, Nicolas Papernot, Ryan Rogers, Milan Shen, Shuang Song, Weijie Su, Andreas Terzis, Abhradeep Thakurta, Sergei Vassilvitskii, Yu-Xiang Wang, Li~Xiong, Sergey Yekhanin, Da~Yu, Huanyu Zhang, and Wanrong Zhang. 2024.
\newblock \href {https://hdsr.mitpress.mit.edu/pub/sl9we8gh} {{Advancing Differential Privacy: Where We Are Now and Future Directions for Real-World Deployment}}.
\newblock \emph{Harvard Data Science Review}, 6(1).

\bibitem[{Devlin et~al.(2019)Devlin, Chang, Lee, and Toutanova}]{devlin-etal-2019-bert}
Jacob Devlin, Ming-Wei Chang, Kenton Lee, and Kristina Toutanova. 2019.
\newblock \href {https://doi.org/10.18653/v1/N19-1423} {{BERT}: Pre-training of deep bidirectional transformers for language understanding}.
\newblock In \emph{Proceedings of the 2019 Conference of the North {A}merican Chapter of the Association for Computational Linguistics: Human Language Technologies, Volume 1 (Long and Short Papers)}, pages 4171--4186, Minneapolis, Minnesota. Association for Computational Linguistics.

\bibitem[{Dockhorn et~al.(2023)Dockhorn, Cao, Vahdat, and Kreis}]{dockhorn2023differentially}
Tim Dockhorn, Tianshi Cao, Arash Vahdat, and Karsten Kreis. 2023.
\newblock \href {http://arxiv.org/abs/2210.09929} {Differentially private diffusion models}.

\bibitem[{Dwork(2006)}]{10.1007/11787006_1}
Cynthia Dwork. 2006.
\newblock {Differential Privacy}.
\newblock In \emph{Automata, Languages and Programming}, pages 1--12, Berlin, Heidelberg. Springer Berlin Heidelberg.

\bibitem[{Dwork et~al.(2006)Dwork, McSherry, Nissim, and Smith}]{10.1007/11681878_14}
Cynthia Dwork, Frank McSherry, Kobbi Nissim, and Adam Smith. 2006.
\newblock {Calibrating Noise to Sensitivity in Private Data Analysis}.
\newblock In \emph{Theory of Cryptography}, pages 265--284, Berlin, Heidelberg. Springer Berlin Heidelberg.

\bibitem[{Dwork and Roth(2014)}]{10.1561/0400000042}
Cynthia Dwork and Aaron Roth. 2014.
\newblock \href {https://doi.org/10.1561/0400000042} {{The Algorithmic Foundations of Differential Privacy}}.
\newblock \emph{Found. Trends Theor. Comput. Sci.}, 9(3–4):211–407.

\bibitem[{Fereidouni et~al.(2022)Fereidouni, Mosharrof, Farooq, and Siddique}]{10020586}
Moghis Fereidouni, Adib Mosharrof, Umar Farooq, and A.B. Siddique. 2022.
\newblock \href {https://doi.org/10.1109/BigData55660.2022.10020586} {{Proactive Prioritization of App Issues via Contrastive Learning}}.
\newblock In \emph{2022 IEEE International Conference on Big Data (Big Data)}, pages 535--544.

\bibitem[{Ghalebikesabi et~al.(2023)Ghalebikesabi, Berrada, Gowal, Ktena, Stanforth, Hayes, De, Smith, Wiles, and Balle}]{ghalebikesabi2023differentially}
Sahra Ghalebikesabi, Leonard Berrada, Sven Gowal, Ira Ktena, Robert Stanforth, Jamie Hayes, Soham De, Samuel~L. Smith, Olivia Wiles, and Borja Balle. 2023.
\newblock \href {https://doi.org/10.48550/arXiv.2302.13861} {{Differentially Private Diffusion Models Generate Useful Synthetic Images}}.
\newblock \emph{arXiv preprint}.

\bibitem[{Gong et~al.(2023{\natexlab{a}})Gong, Li, Feng, Wu, and Kong}]{gong2022diffuseq}
Shansan Gong, Mukai Li, Jiangtao Feng, Zhiyong Wu, and Lingpeng Kong. 2023{\natexlab{a}}.
\newblock {DiffuSeq: Sequence to Sequence Text Generation with Diffusion Models}.
\newblock In \emph{International Conference on Learning Representations, ICLR}.

\bibitem[{Gong et~al.(2023{\natexlab{b}})Gong, Li, Feng, Wu, and Kong}]{gong-etal-2023-diffuseq}
Shansan Gong, Mukai Li, Jiangtao Feng, Zhiyong Wu, and Lingpeng Kong. 2023{\natexlab{b}}.
\newblock \href {https://doi.org/10.18653/v1/2023.findings-emnlp.660} {{D}iffu{S}eq-v2: Bridging discrete and continuous text spaces for accelerated {S}eq2{S}eq diffusion models}.
\newblock In \emph{Findings of the Association for Computational Linguistics: EMNLP 2023}, pages 9868--9875, Singapore. Association for Computational Linguistics.

\bibitem[{Habernal et~al.(2023)Habernal, Mireshghallah, Thaine, Ghanavati, and Feyisetan}]{habernal-etal-2023-privacy}
Ivan Habernal, Fatemehsadat Mireshghallah, Patricia Thaine, Sepideh Ghanavati, and Oluwaseyi Feyisetan. 2023.
\newblock \href {https://doi.org/10.18653/v1/2023.eacl-tutorials.6} {{Privacy-Preserving Natural Language Processing}}.
\newblock In \emph{Proceedings of the 17th Conference of the European Chapter of the Association for Computational Linguistics: Tutorial Abstracts}, pages 27--30, Dubrovnik, Croatia. Association for Computational Linguistics.

\bibitem[{Harode(2020)}]{Harode2020}
Rohan Harode. 2020.
\newblock \href {https://medium.com/sfu-cspmp/draw-drug-review-analysis-work-96212ed98941} {Draw - drug review analysis work}.

\bibitem[{Hernandez et~al.(2022)Hernandez, Epelde, Alberdi, Cilla, and Rankin}]{HERNANDEZ202228}
Mikel Hernandez, Gorka Epelde, Ane Alberdi, Rodrigo Cilla, and Debbie Rankin. 2022.
\newblock \href {https://doi.org/https://doi.org/10.1016/j.neucom.2022.04.053} {{Synthetic data generation for tabular health records: A systematic review}}.
\newblock \emph{Neurocomputing}, 493:28--45.

\bibitem[{Ho et~al.(2020)Ho, Jain, and Abbeel}]{NEURIPS2020_4c5bcfec}
Jonathan Ho, Ajay Jain, and Pieter Abbeel. 2020.
\newblock \href {https://proceedings.neurips.cc/paper_files/paper/2020/file/4c5bcfec8584af0d967f1ab10179ca4b-Paper.pdf} {{Denoising Diffusion Probabilistic Models}}.
\newblock In \emph{Advances in Neural Information Processing Systems}, volume~33, pages 6840--6851. Curran Associates, Inc.

\bibitem[{Igamberdiev et~al.(2022)Igamberdiev, Arnold, and Habernal}]{igamberdiev-etal-2022-dp}
Timour Igamberdiev, Thomas Arnold, and Ivan Habernal. 2022.
\newblock \href {https://aclanthology.org/2022.coling-1.258} {{DP}-rewrite: Towards reproducibility and transparency in differentially private text rewriting}.
\newblock In \emph{Proceedings of the 29th International Conference on Computational Linguistics}, pages 2927--2933, Gyeongju, Republic of Korea. International Committee on Computational Linguistics.

\bibitem[{Ji et~al.(2021)Ji, Li, Huang, and Cambria}]{Ji2021}
Shaoxiong Ji, Xue Li, Zi~Huang, and Erik Cambria. 2021.
\newblock \href {https://doi.org/10.1007/s00521-021-06208-y} {Suicidal ideation and mental disorder detection with attentive relation networks}.
\newblock \emph{Neural Computing and Applications}, 34(13):10309–10319.

\bibitem[{Klimt and Yang(2004)}]{enron}
Bryan Klimt and Yiming Yang. 2004.
\newblock \href {https://doi.org/10.1007/978-3-540-30115-8_22} {{The Enron Corpus: A New Dataset for Email Classification Research}}.
\newblock In \emph{Lecture Notes in Artificial Intelligence (Subseries of Lecture Notes in Computer Science)}, volume 3201, pages 217--226.

\bibitem[{Le~Scao~et al.(2023)}]{workshop2023bloom}
Teven Le~Scao~et al. 2023.
\newblock \href {https://doi.org/10.48550/arXiv.2211.05100} {{BLOOM: A 176B-Parameter Open-Access Multilingual Language Model}}.
\newblock \emph{arXiv preprint}.

\bibitem[{LeCun et~al.(2010)LeCun, Cortes, and Burges}]{lecun2010mnist}
Yann LeCun, Corinna Cortes, and CJ~Burges. 2010.
\newblock {MNIST handwritten digit database}.
\newblock \emph{ATT Labs [Online]. Available: http://yann.lecun.com/exdb/mnist}, 2.

\bibitem[{Lewis et~al.(2020)Lewis, Liu, Goyal, Ghazvininejad, Mohamed, Levy, Stoyanov, and Zettlemoyer}]{lewis-etal-2020-bart}
Mike Lewis, Yinhan Liu, Naman Goyal, Marjan Ghazvininejad, Abdelrahman Mohamed, Omer Levy, Veselin Stoyanov, and Luke Zettlemoyer. 2020.
\newblock \href {https://doi.org/10.18653/v1/2020.acl-main.703} {{BART}: Denoising sequence-to-sequence pre-training for natural language generation, translation, and comprehension}.
\newblock In \emph{Proceedings of the 58th Annual Meeting of the Association for Computational Linguistics}, pages 7871--7880, Online. Association for Computational Linguistics.

\bibitem[{Li et~al.(2012)Li, Qardaji, and Su}]{Li.et.al.2012.ASIACCS}
Ninghui Li, Wahbeh Qardaji, and Dong Su. 2012.
\newblock \href {https://doi.org/10.1145/2414456.2414474} {On {{Sampling}}, {{Anonymization}}, and {{Differential Privacy Or}}, {{K-Anonymization Meets Differential Privacy}}}.
\newblock In \emph{Proceedings of the 7th {{ACM Symposium}} on {{Information}}, {{Computer}} and {{Communications Security}}}, pages 32--33. {ACM}.

\bibitem[{Li et~al.(2022)Li, Thickstun, Gulrajani, Liang, and Hashimoto}]{li2022diffusionlm}
Xiang Li, John Thickstun, Ishaan Gulrajani, Percy~S Liang, and Tatsunori~B Hashimoto. 2022.
\newblock {Diffusion-LM Improves Controllable Text Generation}.
\newblock In \emph{Advances in Neural Information Processing Systems}, volume~35, pages 4328--4343. Curran Associates, Inc.

\bibitem[{Li et~al.(2023)Li, Bubeck, Eldan, Giorno, Gunasekar, and Lee}]{li2023textbooksneediiphi15}
Yuanzhi Li, Sébastien Bubeck, Ronen Eldan, Allie~Del Giorno, Suriya Gunasekar, and Yin~Tat Lee. 2023.
\newblock \href {http://arxiv.org/abs/2309.05463} {Textbooks are all you need ii: phi-1.5 technical report}.

\bibitem[{Lin et~al.(2023)Lin, Gong, Shen, Wu, Fan, Lin, Duan, and Chen}]{10.5555/3618408.3619275}
Zhenghao Lin, Yeyun Gong, Yelong Shen, Tong Wu, Zhihao Fan, Chen Lin, Nan Duan, and Weizhu Chen. 2023.
\newblock {Text generation with diffusion language models: a pre-training approach with continuous paragraph denoise}.
\newblock In \emph{Proceedings of the 40th International Conference on Machine Learning}, ICML'23. JMLR.org.

\bibitem[{Luo(2022)}]{luo2022understanding}
Calvin Luo. 2022.
\newblock \href {https://doi.org/10.48550/arXiv.2208.11970} {{Understanding Diffusion Models: A Unified Perspective}}.
\newblock \emph{arXiv preprint}.

\bibitem[{Maas et~al.(2011)Maas, Daly, Pham, Huang, Ng, and Potts}]{maas-etal-2011-learning}
Andrew~L. Maas, Raymond~E. Daly, Peter~T. Pham, Dan Huang, Andrew~Y. Ng, and Christopher Potts. 2011.
\newblock \href {https://aclanthology.org/P11-1015} {Learning word vectors for sentiment analysis}.
\newblock In \emph{Proceedings of the 49th Annual Meeting of the Association for Computational Linguistics: Human Language Technologies}, pages 142--150, Portland, Oregon, USA. Association for Computational Linguistics.

\bibitem[{Mattern et~al.(2022)Mattern, Jin, Weggenmann, Schoelkopf, and Sachan}]{mattern-etal-2022-differentially}
Justus Mattern, Zhijing Jin, Benjamin Weggenmann, Bernhard Schoelkopf, and Mrinmaya Sachan. 2022.
\newblock \href {https://doi.org/10.18653/v1/2022.emnlp-main.323} {{Differentially Private Language Models for Secure Data Sharing}}.
\newblock In \emph{Proceedings of the 2022 Conference on Empirical Methods in Natural Language Processing}, pages 4860--4873, Abu Dhabi, United Arab Emirates. Association for Computational Linguistics.

\bibitem[{Narayan et~al.(2018)Narayan, Cohen, and Lapata}]{narayan-etal-2018-dont}
Shashi Narayan, Shay~B. Cohen, and Mirella Lapata. 2018.
\newblock \href {https://doi.org/10.18653/v1/D18-1206} {Don{'}t give me the details, just the summary! topic-aware convolutional neural networks for extreme summarization}.
\newblock In \emph{Proceedings of the 2018 Conference on Empirical Methods in Natural Language Processing}, pages 1797--1807, Brussels, Belgium. Association for Computational Linguistics.

\bibitem[{Nasr et~al.(2023)Nasr, Carlini, Hayase, Jagielski, Cooper, Ippolito, Choquette-Choo, Wallace, Tramèr, and Lee}]{nasr2023scalable}
Milad Nasr, Nicholas Carlini, Jonathan Hayase, Matthew Jagielski, A.~Feder Cooper, Daphne Ippolito, Christopher~A. Choquette-Choo, Eric Wallace, Florian Tramèr, and Katherine Lee. 2023.
\newblock \href {https://doi.org/10.48550/arXiv.2311.17035} {{Scalable Extraction of Training Data from (Production) Language Models}}.
\newblock \emph{arXiv preprint}.

\bibitem[{Piktus et~al.(2023)Piktus, Akiki, Villegas, Lauren{\c{c}}on, Dupont, Luccioni, Jernite, and Rogers}]{piktus-etal-2023-roots}
Aleksandra Piktus, Christopher Akiki, Paulo Villegas, Hugo Lauren{\c{c}}on, G{\'e}rard Dupont, Sasha Luccioni, Yacine Jernite, and Anna Rogers. 2023.
\newblock \href {https://doi.org/10.18653/v1/2023.acl-demo.29} {The {ROOTS} search tool: Data transparency for {LLM}s}.
\newblock In \emph{Proceedings of the 61st Annual Meeting of the Association for Computational Linguistics (Volume 3: System Demonstrations)}, pages 304--314, Toronto, Canada. Association for Computational Linguistics.

\bibitem[{Radford et~al.(2019)Radford, Wu, Child, Luan, Amodei, and Sutskever}]{radford2019language}
Alec Radford, Jeffrey Wu, Rewon Child, David Luan, Dario Amodei, and Ilya Sutskever. 2019.
\newblock \href {https://cdn.openai.com/better-language-models/language_models_are_unsupervised_multitask_learners.pdf} {{Language Models are Unsupervised Multitask Learners}}.

\bibitem[{Raffel et~al.(2020)Raffel, Shazeer, Roberts, Lee, Narang, Matena, Zhou, Li, and Liu}]{10.5555/3455716.3455856}
Colin Raffel, Noam Shazeer, Adam Roberts, Katherine Lee, Sharan Narang, Michael Matena, Yanqi Zhou, Wei Li, and Peter~J. Liu. 2020.
\newblock {Exploring the Limits of Transfer Learning with a Unified Text-to-Text Transformer}.
\newblock \emph{Journal of Machine Learning Research}, 21(140):1--67.

\bibitem[{Ramesh et~al.(2022)Ramesh, Dhariwal, Nichol, Chu, and Chen}]{ramesh2022hierarchical}
Aditya Ramesh, Prafulla Dhariwal, Alex Nichol, Casey Chu, and Mark Chen. 2022.
\newblock {Hierarchical text-conditional image generation with clip latents}.
\newblock \emph{arXiv preprint arXiv:2204.06125}, 1(2):3.

\bibitem[{Stadler et~al.(2022)Stadler, Oprisanu, and Troncoso}]{277172}
Theresa Stadler, Bristena Oprisanu, and Carmela Troncoso. 2022.
\newblock \href {https://www.usenix.org/conference/usenixsecurity22/presentation/stadler} {{Synthetic Data --Anonymisation Groundhog Day}}.
\newblock In \emph{31st USENIX Security Symposium (USENIX Security 22)}, pages 1451--1468, Boston, MA. USENIX Association.

\bibitem[{Vaswani et~al.(2017)Vaswani, Shazeer, Parmar, Uszkoreit, Jones, Gomez, Kaiser, and Polosukhin}]{NIPS2017_3f5ee243}
Ashish Vaswani, Noam Shazeer, Niki Parmar, Jakob Uszkoreit, Llion Jones, Aidan~N Gomez, \L~ukasz Kaiser, and Illia Polosukhin. 2017.
\newblock \href {https://proceedings.neurips.cc/paper_files/paper/2017/file/3f5ee243547dee91fbd053c1c4a845aa-Paper.pdf} {{Attention is All you Need}}.
\newblock In \emph{Advances in Neural Information Processing Systems}, volume~30. Curran Associates, Inc.

\bibitem[{Wolf et~al.(2020)Wolf, Debut, Sanh, Chaumond, Delangue, Moi, Cistac, Rault, Louf, Funtowicz, Davison, Shleifer, von Platen, Ma, Jernite, Plu, Xu, Le~Scao, Gugger, Drame, Lhoest, and Rush}]{wolf-etal-2020-transformers}
Thomas Wolf, Lysandre Debut, Victor Sanh, Julien Chaumond, Clement Delangue, Anthony Moi, Pierric Cistac, Tim Rault, Remi Louf, Morgan Funtowicz, Joe Davison, Sam Shleifer, Patrick von Platen, Clara Ma, Yacine Jernite, Julien Plu, Canwen Xu, Teven Le~Scao, Sylvain Gugger, Mariama Drame, Quentin Lhoest, and Alexander Rush. 2020.
\newblock \href {https://doi.org/10.18653/v1/2020.emnlp-demos.6} {Transformers: State-of-the-art natural language processing}.
\newblock In \emph{Proceedings of the 2020 Conference on Empirical Methods in Natural Language Processing: System Demonstrations}, pages 38--45, Online. Association for Computational Linguistics.

\bibitem[{Wood et~al.(2018)Wood, Altman, Bembenek, Bun, Gaboardi, Honaker, Nissim, O'Brien, Steinke, and Vadhan}]{wood2018differential}
Alexandra Wood, Micah Altman, Aaron Bembenek, Mark Bun, Marco Gaboardi, James Honaker, Kobbi Nissim, David~R O'Brien, Thomas Steinke, and Salil Vadhan. 2018.
\newblock {Differential privacy: A primer for a non-technical audience}.
\newblock \emph{Vand. J. Ent. \& Tech. L.}, 21:209.

\bibitem[{Yang et~al.(2023)Yang, Zhang, Song, Hong, Xu, Zhao, Zhang, Cui, and Yang}]{10.1145/3626235}
Ling Yang, Zhilong Zhang, Yang Song, Shenda Hong, Runsheng Xu, Yue Zhao, Wentao Zhang, Bin Cui, and Ming-Hsuan Yang. 2023.
\newblock \href {https://doi.org/10.1145/3626235} {{Diffusion Models: A Comprehensive Survey of Methods and Applications}}.
\newblock \emph{ACM Comput. Surv.}, 56(4).

\bibitem[{Yousefpour et~al.(2021)Yousefpour, Shilov, Sablayrolles, Testuggine, Prasad, Malek, Nguyen, Ghosh, Bharadwaj, Zhao, Cormode, and Mironov}]{opacus}
Ashkan Yousefpour, Igor Shilov, Alexandre Sablayrolles, Davide Testuggine, Karthik Prasad, Mani Malek, John Nguyen, Sayan Ghosh, Akash Bharadwaj, Jessica Zhao, Graham Cormode, and Ilya Mironov. 2021.
\newblock {Opacus: User-Friendly Differential Privacy Library in PyTorch}.
\newblock \emph{arXiv preprint arXiv:2109.12298}.

\bibitem[{Yuan et~al.(2024)Yuan, Yuan, Tan, Huang, and Huang}]{yuan-etal-2024-text}
Hongyi Yuan, Zheng Yuan, Chuanqi Tan, Fei Huang, and Songfang Huang. 2024.
\newblock \href {https://doi.org/10.18653/v1/2024.naacl-long.2} {Text diffusion model with encoder-decoder transformers for sequence-to-sequence generation}.
\newblock In \emph{Proceedings of the 2024 Conference of the North American Chapter of the Association for Computational Linguistics: Human Language Technologies (Volume 1: Long Papers)}, pages 22--39, Mexico City, Mexico. Association for Computational Linguistics.

\bibitem[{Yue et~al.(2023)Yue, Inan, Li, Kumar, McAnallen, Shajari, Sun, Levitan, and Sim}]{yue-etal-2023-synthetic}
Xiang Yue, Huseyin Inan, Xuechen Li, Girish Kumar, Julia McAnallen, Hoda Shajari, Huan Sun, David Levitan, and Robert Sim. 2023.
\newblock \href {https://doi.org/10.18653/v1/2023.acl-long.74} {{Synthetic Text Generation with Differential Privacy: A Simple and Practical Recipe}}.
\newblock In \emph{Proceedings of the 61st Annual Meeting of the Association for Computational Linguistics (Volume 1: Long Papers)}, pages 1321--1342, Toronto, Canada. Association for Computational Linguistics.

\end{thebibliography}
\bibliographystyle{acl_natbib}

\appendix

\section{Hyperparameter Settings}
\label{sec:hyperparams}
\begin{table*}
\centering
\begin{small}
\begin{tabular}{p{2cm}rrrrrrr} \toprule
& \multicolumn{3}{|c|}{DP} & \multicolumn{3}{c}{non-DP} \\
& \multicolumn{1}{|c}{Epoch*/Steps} & BSZ* & \multicolumn{1}{c|}{LR} & Epoch/Steps & BSZ & LR \\ \midrule
\multicolumn{1}{l|}{BART} & 8 & 32 & \multicolumn{1}{c|}{2e-5} & 4 & 32 & 2e-5 \\
\multicolumn{1}{l|}{BLOOM}  & 8 & 16 & \multicolumn{1}{c|}{2e-5} & 4 & 16 & 2e-5 \\
\multicolumn{1}{l|}{PHI}   & 8 & 8 & \multicolumn{1}{c|}{2e-5} & 4 & 8 & 2e-5 \\ \midrule
\multicolumn{1}{l|}{DIFFUSEQ} & 200,000 & 128 & \multicolumn{1}{c|}{1e-4}& 100,000 & 128 & 1e-4 \\
\multicolumn{1}{l|}{SEQDIFFUSEQ}   & 400,000 & 128 & \multicolumn{1}{c|}{1e-4} & 200,000 & 128 & 1e-4 \\
\multicolumn{1}{l|}{GENIE}   & 50,000 & 256 & \multicolumn{1}{c|}{1e-5} & 50,000 & 128 & 50000 \\
\end{tabular}
\end{small}
\caption{Hyperparameter settings for all datasets and (non-) privacy setting (\textit{BSZ = batch size, LR = learning rate}). Epoch* in DP describes the number of steps taken in the usual non-privacy setting, even though the data loader performs random sampling with replacement. BSZ*: During DP training, the data loader also performs poisson-sampling on the batch size, so that each batch is possibly different in size for each training step. We display the number of epochs for the baselines and number of training steps for the text diffusion models}
    \label{tab:hypers}
\end{table*}

We present the hyperparameter settings of our experiments in Table \ref{tab:hypers}.

\section{Synthetic Texts and Classifier Performance}
\label{moresamples}
\begin{table}[ht]
\centering
\begin{small}
\begin{tabular}{p{2cm}rrr} \toprule
\multicolumn{1}{l|}{Dataset} & Acc & MF1 & PPL\\ \midrule
\multicolumn{1}{l|}{SPAM} & 0.98 & 0.98 & 475\\
\multicolumn{1}{l|}{SWMH}  & 0.63 & 0.63 & 234 \\
\multicolumn{1}{l|}{THUMBSUP}   & 0.39 & 0.39 & 3e+16 \\
\multicolumn{1}{l|}{WEBMD}   & 0.39 & 0.39 & 4e+13 \\
\end{tabular}
\end{small}
\caption{\label{bresults} Accuracy (Acc) and macro F1-score (MF1) results of the BERT classifier trained on a random subsample of the original training data and tested on the original test sets, as well as the average perplexity (PPL) of the subsample.}
\end{table}
As the metric scores for the synthetic data, especially for the Thumbs-Up and WebMD dataset are close to random chance, we also report the results we achieved by training a BERT classifier on a balanced, random subsample of the original training data, evaluated on the test set, in Table \ref{bresults}.

\begin{figure}[t!]
    \centering
    \begin{framed}
    \fontsize{9pt}{12pt}\selectfont  

    \begin{description}
    \item[\textbf{Instruction:}] write a mild app review: 
    \item[\textbf{BART}] i love this game but i can't play it
    \item[\textbf{BLOOM}] slop like for a game. im willing to pay to play. but i'm so dumb in all way to play it. u want to get back coins? i'm dumb enough to let it go. i'd rather want to play a game and still be good since nis won't fix and won't fix me...
    \item[\textbf{PHI-1.5}] Since Kevin was on a strict diet for the past month, he ate his usual ice cream without realizing that it was loaded with sugar and empty calories. Once he finally broke this habit, he felt great, but couldn't stop thinking about the old unhealthy relationship habits that kept him captive...
    \item[\textbf{DiffuSeq}] ??????????????????????????????????
    ??????????????????????????????????????
    ????????????????????????????????????...
    \item[\textbf{SeqDiffuSeq}] . work for there though.
    \item[\textbf{GENIE}], we put up a game game and the rest of the game.
    \end{description}
	\end{framed}
    \caption{\label{fig:e3}Exemplary synthetic texts generated by models trained on Thumbs Up with $\varepsilon = 3$. Some texts are truncated to fit into the scope of this paper, indicated "..." at the end.} 
\end{figure}

\begin{figure}[t!]

    \centering
    \begin{framed}
    \fontsize{9pt}{12pt}\selectfont  

    \begin{description}
    \item[\textbf{Instruction:}] write a mild app review: 
    \item[\textbf{BART}] good game, but it doesn't work very well. i've been playing for a few months now and every time i try to start the game, it won't load.
    \item[\textbf{BLOOM}] ive lost my payment when my game was out but if an app used to be available then then i now have lost it for this type of game. now the app will return to my bank account and money but i i lost my money but it did return. i can get to my apps though with the issue...
    \item[\textbf{PHI-1.5}] ! i am happy you like this game.        Two co-workers, Tom and Lisa, have been working together at a clothing store for several months. Last week, their boss announced a company-sponsored event that involved team building and outdoor activities...
    \item[\textbf{DiffuSeq}] nguyen nguyen nguyen nguyen nguyen nguyen nguyen nguyen nguyen nguyen nguyen nguyen nguyen nguyen nguyen nguyen nguyen nguyen nguyen nguyen nguyen nguyen ...
    \item[\textbf{SeqDiffuSeq}] it's true.
    \item[\textbf{GENIE}] to have to argue that it's going to be happy off, and so
    \end{description}
	\end{framed}
    \caption{\label{fig:e8}Exemplary synthetic texts generated by models trained on Thumbs Up with $\varepsilon = 8$. Some texts are truncated to fit into the scope of this paper, indicated "..." at the end.} 
\end{figure}
\begin{figure}[t!]
    \centering
    \begin{framed}
    \fontsize{9pt}{12pt}\selectfont  

    \begin{description}
    \item[\textbf{Instruction:}] write a mild app review: 
    \item[\textbf{BART}] i love this game. i've been playing it for over a year now. i used to love it. but lately it's been so laggy. i don't know if it's my phone or the game but it's not my internet connection. i'm on a new phone and i can't even get into the game anymore...
    \item[\textbf{BLOOM}] ive been having a problem where this app randomly freezes and i cant access my stuff. i've been really mad about this but not sure why. i keep going through this process to get my rewards and now it's finished!...
    \item[\textbf{PHI-1.5}] ive started this game before and i got to level 30 or 31 but when i uninstalled it and reinstalled it i had to start from level one, it deleted all of my progress please fix this ive played this game every day for like 2 years and just recently played it for the first time i reinstled it...
    \item[\textbf{DiffuSeq}] i can't get into the game. uninstall and reinstall the game, i'm not anymore.
    \item[\textbf{SeqDiffuSeq}] this game is trash. deleted it to download and played after 3 hours. not worth it. 
    \item[\textbf{GENIE}] this one is the most worst game. i ever did
    \end{description}
	\end{framed}
    \caption{\label{fig:inf}Exemplary synthetic texts generated by models trained on Thumbs Up with $\varepsilon = \infty$. Some texts are truncated to fit into the scope of this paper, indicated "..." at the end.} 
\end{figure}

We present synthetic texts generated from models trained on the Thumbs-Up dataset, across all privacy budgets in Figures \ref{fig:e3}, \ref{fig:e8} and \ref{fig:inf}.

\section{Dataset details}
\label{data-dist}
\begin{table*}[!t]
\centering
\begin{small}
\begin{tabular}{p{2cm}rrrrr} \toprule
 \textbf{SPAM}    & non-spam & spam \\ \midrule
 Train & 19,043    & 23,132 \\
 Test  & 4,761     & 5,783 \\ \toprule \toprule
 \textbf{SWMH}    & anxiety & bipolar & depression & offmychest & suicidewatch\\ \midrule
 Train & 6,136    & 4,932 & 11,940 & 6,550 & 5,265 \\
 Test  & 1,911 & 1492 & 3,774 & 1,687 & 2,018 \\ \toprule \toprule
 \textbf{THUMBSUP}    & mild & notable & concerning & serious & hot \\ \midrule
 Train & 30,933    & 30,933 & 30,933 & 30,933 & 30,933 \\
 Test  & 403,534 & 110,172 & 33,282 & 15,474 & 7,862 \\ \toprule \toprule
 \textbf{WEBMD}    & terrible & poor & neutral & good & great \\ \midrule
 Train & 28273    & 28273 & 28274 & 28274 & 28274 \\
 Test  & 7069 & 7068 & 7069 & 7068 & 7068 \\
\end{tabular}
\end{small}
\caption{\label{tab:data-dist} Data distribution of our processed datasets}
\end{table*}

We display the number of samples per label in each of our processed data splits in Table \ref{tab:data-dist}. 

\section{DP-SGD implementation details}
\label{dp-sgd-section}
While the Opacus library almost supports all standard layers by the PyTorch \cite{pytorch} library, it often has conflicts with custom layers from transformer-based models, especially positional embedding layers.
We therefore detail in this section our workarounds for training our LLMs and diffusion models when unsupported layers are encountered.
\paragraph{BART}
In our experiments, we utilize the conditional text generation version of BART provided by the transformers \cite{wolf-etal-2020-transformers} library.
As all layers except the positional embedding layer of BARTs decoder are compatible with Opacus, we decide to freeze the positional embeddings during DP-SGD fine-tuning, which allows Opacus to correctly calculate the gradients for all other layers, while leaving the positional embeddings in its pretrained state. The metric results of the SPAM dataset in Table \ref{expresults} show that this does not negatively impact our experimental results. 
\paragraph{BLOOM and Phi-1.5}
For Phi-1.5 and BLOOM, we use the causal language models from the transformers library, which do not conflict with Opacus, so no change was necessary to train both models with DP.
\paragraph{DiffuSeq}
DiffuSeq also relies on a custom positional embedding. While Opacus is able to capture the gradient of this layer, it does not correctly expand it along the first dimension to the current batch (lot) size. 
Manually expanding the gradient of the positional embedding weights to the current batch (lot) size, enables DP-SGD training with DiffuSeq, as all other layers are compatible. 
\paragraph{SeqDiffuSeq}
As SeqDiffuSeq is built upon the BART model from the transformers library, the same method we applied to BART works here: Simply freezing the weights of the positional embedding layer does the trick.
\paragraph{GENIE}
Similar to SeqDiffuSeq, GENIE contains two positional embedding layers where the gradient is captured by Opacus, but not correctly expanded.
We therefore again expand the two positional embedding weight gradients to the batch (lot) size, which solves the incompability issues with Opacus.
\FloatBarrier
\begin{table*}[t!]
\begin{small}
\begin{center}
\begin{tabular}{p{2cm}rrrrrr} \toprule
\textbf{DRUGS} & \multicolumn{2}{c}{$\varepsilon = 3$} & \multicolumn{2}{c}{$\varepsilon = 8$} & \multicolumn{2}{c}{$\varepsilon = \infty$} \\
n = 1000 & Acc & MF1 & Acc & MF1 & Acc & MF1 \\ \midrule
\multicolumn{1}{l|}{BART} & 0.56$\pm$0.13 & \multicolumn{1}{c|}{\textbf{0.54$\pm$0.15}} & \textbf{0.61$\pm$0.03}  & \multicolumn{1}{c|}{\textbf{0.61$\pm$0.03}} & 0.69+$\pm$0.09 &  \multicolumn{1}{c}{0.69$\pm$0.09}  \\
\multicolumn{1}{l|}{PHI-1.5}   & \textbf{0.57$\pm$0.10} & \multicolumn{1}{c|}{0.42$\pm$0.05} & 0.46$\pm$0.12 & \multicolumn{1}{c|}{0.41$\pm$0.12} & 0.73$\pm$0.05 & \multicolumn{1}{c}{0.61$\pm$0.11}  \\ \midrule
\multicolumn{1}{l|}{GENIE}   & 0.44$\pm$0.01 & \multicolumn{1}{c|}{0.43$\pm$0.02} & 0.46$\pm$0.02  & \multicolumn{1}{c|}{.45$\pm$0.02} & \textbf{0.79$\pm$0.02} & \multicolumn{1}{c}{\textbf{0.75$\pm$0.05}}  \\ \toprule
\textbf{DRUGS} & \multicolumn{2}{c}{$\varepsilon = 3$} & \multicolumn{2}{c}{$\varepsilon = 8$} & \multicolumn{2}{c}{$\varepsilon = \infty$} \\
n = 5000& Acc & MF1 &  Acc & MF1 & Acc & MF1 \\ \midrule
\multicolumn{1}{l|}{BART} & 0.51$\pm$0.12 & \multicolumn{1}{c|}{0.39$\pm$0.03} & 0.63$\pm$0.06  & \multicolumn{1}{c|}{0.63$\pm$0.06} & 0.73$\pm$0.03 & \multicolumn{1}{c}{0.73$\pm$0.03}  \\
\multicolumn{1}{l|}{PHI-1.5}   & \textbf{0.77$\pm$0.03} & \multicolumn{1}{c|}{\textbf{0.75$\pm$0.02}} & \textbf{0.76$\pm$0.01}  & \multicolumn{1}{c|}{\textbf{0.75$\pm$0.01}} & \textbf{0.79$\pm$0.02} & \multicolumn{1}{c}{0.73$\pm$0.03}  \\ \midrule
\multicolumn{1}{l|}{GENIE}   & 0.45$\pm$0.02 & \multicolumn{1}{c|}{0.44$\pm$0.03} & 0.46$\pm$0.03  & \multicolumn{1}{c|}{0.46$\pm$0.03} & \textbf{0.79$\pm$0.01} & \multicolumn{1}{c}{\textbf{0.78$\pm$0.01}}  \\ \toprule
\textbf{DRUGS} & \multicolumn{2}{c}{$\varepsilon = 3$} & \multicolumn{2}{c}{$\varepsilon = 8$} & \multicolumn{2}{c}{$\varepsilon = \infty$} \\
n = 10000 & Acc & MF1 & Acc & MF1 & Acc & MF1 \\ \midrule
\multicolumn{1}{l|}{BART} & 0.52$\pm$0.12 & \multicolumn{1}{c|}{0.49$\pm$0.14} & 0.65$\pm$0.11  & \multicolumn{1}{c|}{0.60$\pm$0.14} & 0.69$\pm$0.06 & \multicolumn{1}{c}{0.69$\pm$0.06}  \\
\multicolumn{1}{l|}{PHI-1.5}   & \textbf{0.73$\pm$0.05} & \multicolumn{1}{c|}{\textbf{0.67$\pm$0.14}} & \textbf{0.77$\pm$0.02}  & \multicolumn{1}{c|}{\textbf{0.76$\pm$0.02}} & \textbf{0.81$\pm$0.02} &  \multicolumn{1}{c}{0.76$\pm$0.03}  \\ \midrule
\multicolumn{1}{l|}{GENIE}   & 0.45$\pm$0.01 & \multicolumn{1}{c|}{0.45$\pm$0.01} & 0.44$\pm$0.02  & \multicolumn{1}{c|}{0.43$\pm$0.02} & 0.79$\pm$0.03 & \multicolumn{1}{c}{\textbf{0.78$\pm$0.02}}  \\ \toprule
\end{tabular}
\end{center}
\end{small}
\caption{\label{suppresults} Mean ($\pm$ standard deviation) Accuracy (Acc) and macro F1-score (MF1) results of BERT classifiers trained on synthetic data and tested on the original test sets. The best Acc and MF1 values per sample size and privacy budget are highlighted.}
\end{table*}
\FloatBarrier

\section{Supplementary experiments}
\label{supp}

\arr{It turned out to be surprisingly hard to find a publicly available data set that (1) discloses unique authorship information and (2) is not part of the pretraining of any of the models in our main experiments.
Nonetheless, to investigate how text diffusion models and LLMs perform under DP when a one-to-one relation between authors and data points is ensured, we conduct the following supplementary experiment.}

We create a dataset fulfilling both conditions by collecting 50K reviews from the Drugs.com\textregistered \footnote{\url{https://www.drugs.com}} website.
Here, users can write text reviews about medications they took, accompanied by a recommendation score scaling from 1 to 10.
While users can submit anonymous reviews, they can also register a user name which is displayed in their respective review.
We use this feature to uniquify our collected reviews and a dataset of 21K reviews remains. 
We decide to perform binary sentiment analysis on the data by labeling reviews with a score of 5 or below as negative, and reviews with a score above 5 as positive. 
The resulting label distribution is $67\%$ positve and $33\%$ negative.
We also split the data into a train, development and test set (8:1:1) for our experiment.

In contrast to our main experiments, we also measure how the amount of synthetic texts impacts their utility. 
Therefore, we decide to generate 10K samples per model and privacy budget.
We conduct our experiments on BART, PHI-1.5 and GENIE, as parts of Drugs.com\textregistered  may have been in the pretraining data of BLOOM, and generating 10K texts with DiffuSeq and SeqDiffuSeq exceeds our computational budget.
The text classifiers are trained on the synthetic data with five random seeds to improve the stability of the supplementary experiment.

The results are displayed in Table \ref{suppresults}.
We observe that (1) utility metrics generally improve when more samples are introduced, (2) GENIE outperforms BART and PHI-1.5 in terms of MF1 when not finetuned with DP, but falls behind significantly when it was. 
This strengthens our main conclusion that text diffusion models do not outperform LLMs in synthetic text generation under DP constraints.

\end{document}